\documentclass[letterpaper, 10 pt, conference]{ieeeconf}  % Comment this line out if you need a4paper

\IEEEoverridecommandlockouts                              % This command is only needed if 
\usepackage{graphics} % for pdf, bitmapped graphics files
\usepackage{epsfig} % for postscript graphics files
\usepackage{amsmath} % assumes amsmath package installed
\usepackage{amssymb}  % assumes amsmath package installed
\usepackage{mathrsfs}
\usepackage{stmaryrd} % for shortarrow
\usepackage{multirow} % for multirow
\usepackage{makecell}

\usepackage{soul}

\usepackage[pagebackref=true,breaklinks=true,letterpaper=true,colorlinks,bookmarks=false,citecolor=blue,linkcolor=blue]{hyperref}
\usepackage{etoolbox}
\makeatletter
\patchcmd{\@maketitle}{\raggedright}{\centering}{}{}
\patchcmd{\@maketitle}{\raggedright}{\centering}{}{}
\makeatother

\makeatletter
\let\NAT@parse\undefined
\makeatother

\hypersetup{draft} % To fix "\pdfendlink ended up on a different page"

\usepackage{etoolbox}
\makeatletter
\patchcmd{\@makecaption}
  {\scshape}
  {}
  {}
  {}
\makeatletter
\patchcmd{\@makecaption}
  {\\}
  {.\ }
  {}
  {}
\makeatother

\usepackage{caption}

\usepackage{ifthen}
\usepackage{color}
\definecolor{gray}{rgb}{0.5,0.5,0.5}
\definecolor{green}{rgb}{0, 0.4, 0}
\definecolor{orange}{rgb}{1, 0.5, 0}
\definecolor{mahogany}{rgb}{0.75, 0.25, 0.0}
\definecolor{purple}{rgb}{0.6, 0, 0.6}
\definecolor{darkgreen}{rgb}{0, 0.4, 0}
\definecolor{frenchblue}{rgb}{0.0, 0.45, 0.73}
\definecolor{red}{rgb}{1,0,0}
\definecolor{yellow}{rgb}{1,1,0}
\definecolor{magenta}{rgb}{1,0,1}
\definecolor{pink}{rgb}{1,0.412,0.706}
\newboolean{revising}
\setboolean{revising}{false}
\ifthenelse{\boolean{revising}}
{
    \newcommand{\walon}[1]{{\color{red}{#1}}}
    \newcommand{\johnson}[1]{{\color{frenchblue}{#1}}}
	
	\newcommand{\hubert}[1]{{\color{pink}{#1}}}

} {
    \newcommand{\walon}[1]{#1}
    
    \newcommand{\johnson}[1]{#1}
	
	\newcommand{\hubert}[1]{#1}

}
\newcommand{\modelNameIncat}{Input Fusion }
\newcommand{\modelNameIncatPunc}{Input Fusion}
\newcommand{\modelNameCBN}{CCVNorm }
\newcommand{\modelNameCBNPunc}{CCVNorm}
\newcommand{\modelNameHierCBN}{HierCCVNorm }
\newcommand{\modelNameHierCBNPunc}{HierCCVNorm}

\newcommand{\expectation}{\mathop{\mathbb{E}}}

\newcommand{\tabref}{Table~\ref}
\newcommand{\figref}{Fig.~\ref}
\newcommand{\secref}{Sec.~\ref}
\newcommand{\etal}{\textit{et al.}}

\title{\LARGE \bf
3D LiDAR and Stereo Fusion using Stereo Matching Network with Conditional Cost Volume Normalization
}

\begin{document}

\author{Tsun-Hsuan Wang, Hou-Ning Hu, Chieh Hubert Lin, Yi-Hsuan Tsai, Wei-Chen Chiu, Min Sun}
%\author{Tsun-Hsuan Wang$^{1}$, Hou-Ning Hu$^{1}$, Chieh Hubert Lin$^{1}$, Yi-Hsuan Tsai$^{2}$, Wei-Chen Chiu$^{3}$, Min Sun$^{1}$
%\thanks{$^{1}$ National Tsin Hua University}
%\thanks{$^{2}$ NEC Labs America}
%\thanks{$^{3}$ National Chiao Tung University}}

\maketitle
\thispagestyle{empty}
\pagestyle{empty}

%%%%%%%%%%%%%%%%%%%%%%%%%%%%%%%%%%%%%%%%%%%%%%%%%%%%%%%%%%%%%%%%%%%%%%%%%%%%%%%%
\begin{abstract}
The complementary characteristics of active and passive depth sensing techniques motivate the fusion of the LiDAR sensor and stereo camera for improved depth perception. Instead of directly fusing estimated depths across LiDAR and stereo modalities, we take advantages of the stereo matching network with two enhanced techniques: \modelNameIncat and Conditional Cost Volume Normalization (\modelNameCBNPunc) on the LiDAR information. The proposed framework is generic and closely integrated with the cost volume component that is commonly utilized in stereo matching neural networks. We experimentally verify the efficacy and robustness of our method on the KITTI Stereo and Depth Completion datasets, obtaining favorable performance against various fusion strategies. Moreover, we demonstrate that, with a hierarchical extension of \modelNameCBNPunc, the proposed method brings only slight overhead to the stereo matching network in terms of computation time and model size.

\end{abstract}

%%%%%%%%%%%%%%%%%%%%%%%%%%%%%%%%%%%%%%%%%%%%%%%%%%%%%%%%%%%%%%%%%%%%%%%%%%%%%%%%
\section{Introduction}

% 3D perception is important and can be obtained by (1) RGB-D sensor (2) 3D LiDAR (3) passive sensore (stereo)
\walon{
The accurate 3D perception has been desired since its vital role in numerous tasks of robotics and computer vision, such as autonomous driving, localization and mapping, path planning, and 3D reconstruction. Various techniques have been proposed to obtain depth estimation, ranging from active sensing sensors (e.g., RGB-D cameras and 3D LiDAR scanners) to passive sensing ones (e.g., stereo cameras). We observe that these sensors all have their own pros and cons, in which none of them perform well on all practical scenarios. For instance, RGB-D sensor is confined to its short-range depth acquisition and thereby 3D LiDAR is a common alternative in the challenging outdoor environment. However, 3D LiDARs are much more expensive and only provide sparse 3D depth estimates. In contrast, a stereo camera is able to obtain denser depth map based on stereo matching algorithms but is typically incapable of producing reliable matches in regions with repetitive patterns, homogeneous appearance, or large illumination change.} 

\begin{figure}[t!]
\centering
\includegraphics[width=\linewidth]{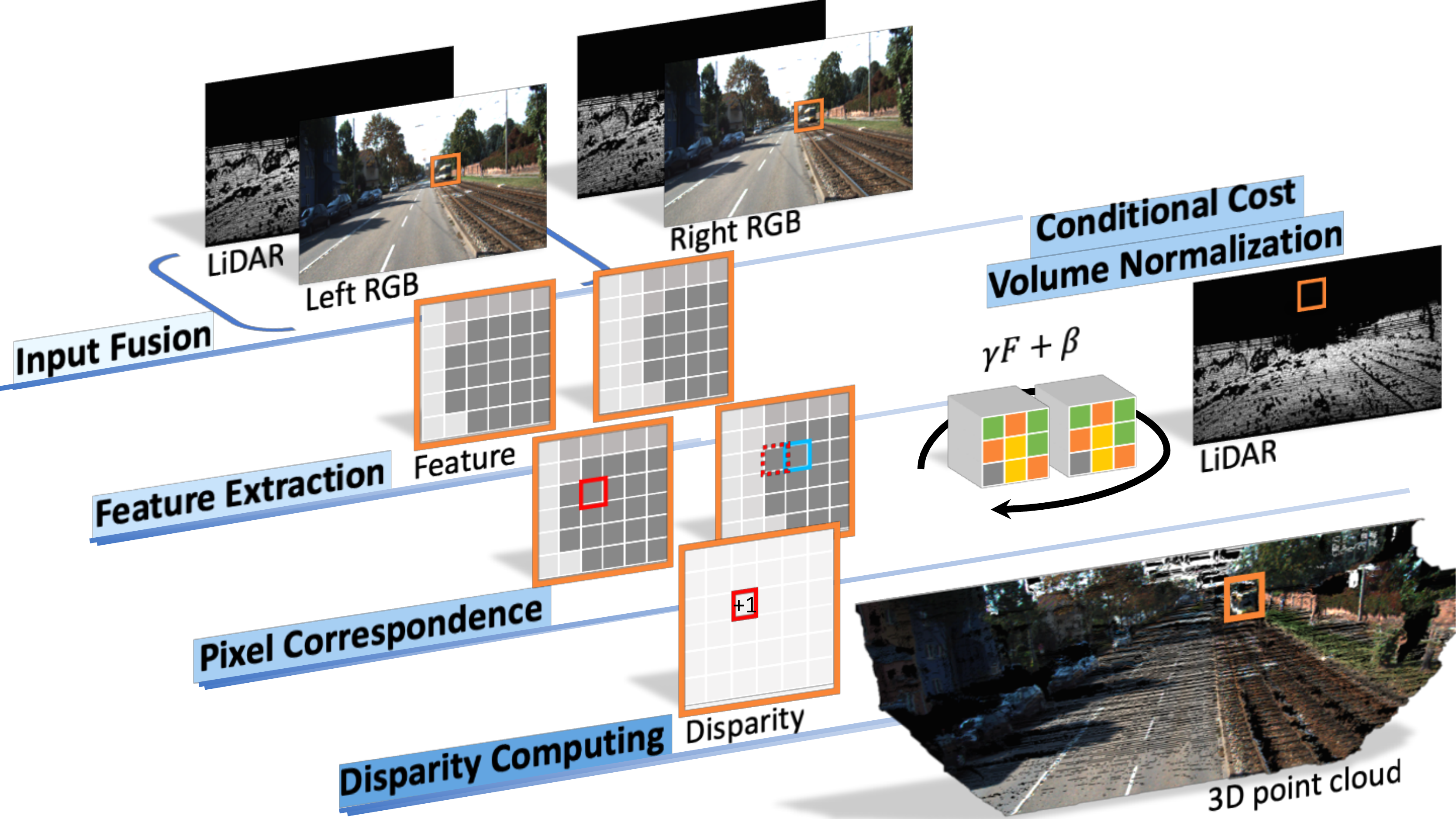}
\caption{\small{Illustration of our method for 3D LiDAR and stereo fusion. The high-level concept of stereo matching pipeline involves 2D feature extraction from the stereo pair, obtaining pixel correspondence, and finally disparity computation. In this paper, we present (1) Input Fusion and (2) Conditional Cost Volume Normalization that are closely integrated with stereo matching networks. By leveraging the complementary nature of LiDAR and stereo modalities, our model produces high-precision disparity estimation.}}
\label{fig:teaser}
\end{figure}

Thanks to the complementary characteristic across different sensors, several works~\cite{LiDARstereoicra18}\cite{LiDARstereoarxiv18} have studied how to fuse multiple modalities in order to provide more accurate and denser depth estimation. In this paper, we consider the fusion of passive stereo camera and active 3D LiDAR sensor, which is a practical and popular choice. \walon{Existing works along this research direction mainly investigate the output-level combination of the dense depth from stereo matching with the sparse measurement from 3D LiDAR.
However, rich information provided in stereo images is thus not well utilized in the procedure of fusion. In order to address this issue, we propose to study the design choices for more closely integrating the 3D LiDAR information into the process of stereo matching methods (illustrated in~\figref{fig:teaser}). The motivation that drives us toward this direction is an observation that typical stereo matching algorithms usually suffer from having ambiguous pixel correspondences across stereo pairs, and thereby 3D LiDAR depth points are able to help reduce the search space of matching and resolve ambiguities.}

\walon{
As depth points from 3D LiDAR sensors are sparse, it is not straightforward to simply treat them as additional features connected to each pixel location of a stereo pair during performing stereo matching.
Instead, we focus on facilitating sparse points to regularize higher-level feature representations in deep learning-based stereo matching.
%
% We hence turn to focus on stereo matching methods which are based on deep-learning models, not only because they are able to outperform traditional ones in a significant margin, but also extract higher-level representations of 3D LiDAR data for integration.
%
Recent state-of-the-arts on deep models of stereo matching are composed of two main components: matching cost computation~\cite{stereomccnn}\cite{stereoefficient} and cost volume regularization~\cite{stereogcnet}\cite{stereocrl}\cite{stereopsmnet}\cite{stereodeepmvs}, where the former basically extracts the deep representation of image patches and the latter builds up the search space to aggregate all potential matches across stereo images with further regularization (e.g., 3D CNN) for predicting the final depth estimate.

Being aligned with these two components, we extend the stereo matching network by proposing two techniques: (1) \textbf{\modelNameIncatPunc} to incorporate the geometric information from sparse LiDAR depth with the RGB images for learning joint feature representations, and (2) \textbf{\modelNameCBNPunc} (\textbf{C}onditional \textbf{C}ost \textbf{V}olume \textbf{Norm}alization) to adaptively regularize cost volume optimization in dependence on LiDAR measurements. It is worth noting that our proposed techniques have little dependency on particular network architectures but only relies on a commonly-utilized cost volume component, thus having more flexibility to be adapted into different models. Extensive experiments are conducted on the KITTI Stereo 2015 Dataset~\cite{kitti2015} and the KITTI Depth Completion Dataset~\cite{kitti2017} to evaluate the effectiveness of our proposed method. In addition, we perform ablation study on different variants of our approach in terms of performance, model size and computation time. Finally, we analyze how our method exploits the additional sparse sensory inputs and provide qualitative comparisons with other fusion schemes to further highlight the strengths and merits of our method.}

\section{Related Works}

\walon{
{\flushleft {\bf Stereo Matching.}} Stereo matching has been a fundamental problem in computer vision. In general, a typical stereo matching algorithm can be summarized into a four-stage pipeline~\cite{stereotaxonomy}, consisting of \textit{matching cost computation}, \textit{cost support aggregation}, \textit{cost volume regularization}, and \textit{disparity refinement}. Even when deep learning is introduced to stereo matching in recent years and brings a significant leap in performance of depth estimation, such design paradigm is still widely utilized. For instance, \cite{stereomccnn} and \cite{stereoefficient} propose to learn a feature representation for matching cost computation by using a deep Siamese network, and then adopt the classical semi-global matching (SGM)~\cite{stereosgm} to refine the disparity map. \cite{sceneflow} and \cite{stereogcnet} further formulate the entire stereo matching pipeline as an end-to-end network, where the cost volume aggregation and regularization are modelled jointly by 3D convolutions. Moreover, \cite{stereocrl} and \cite{stereopsmnet} propose several network designs to better exploit multi-scale and context information. Built upon the powerful learning capacity of deep models, this paper aims to integrate LiDAR information into the procedure of stereo matching networks for a more efficient scheme of fusion.
}

\walon{
{\flushleft {\bf RGB Imagery and LiDAR Fusion.}}
Sensor fusion of RGB imagery and LiDAR data obtain more attention in virtue of its practicability and performance for depth perception. Two different settings are explored by several prior works: LiDAR fused with a monocular image or stereo ones. 
As the depth estimation from a single image is typically based on a regression from pixels, which is inherently unreliable and ambiguous, most of the recent monocular-based works aim to achieve the completion on the sparse depth map obtained by LiDAR sensor with the help of rich information from RGB images\johnson{~\cite{depths2d}\cite{depthsss2d}\cite{depthsicnn}\cite{depthhmsnet}\cite{depthnconv}\cite{depthrgbguide}}, or refine the depth regression by having LiDAR data as a guidance\johnson{~\cite{depthcspn}\cite{depthpnp}}.

On the other hand, since the stereo camera relies on the geometric correspondence across images of different viewing angles, its depth estimates are less ambiguous in terms of the absolute distance between objects in the scene and can be well aligned with the scale of 3D LiDAR measurements. This property of stereo camera makes it a practical choice to be fused with 3D LiDAR data in robotic applications, where the complementary characteristics of passive (stereo) and active (LiDAR) depth sensors are better utilized\johnson{~\cite{LiDARstereo2003}\cite{LiDARstereointegrating}\cite{LiDARstereotof}}. 
% Gandhi \etal~\cite{LiDARstereotof} presents fusion of time-of-flight sensor and stereo camera. 
For instance, Maddern \etal~\cite{LiDARstereoprobfusion} propose a probabilistic framework for fusing LiDAR data with stereo images to generate both the depth and uncertainty estimate. 
%Shivakumar \textit{et al.} suggests combining anisotropic diffusion and semi-global matching to inject range measurements into the cost volume. 
With the power of deep learning, Park \etal~\cite{LiDARstereoicra18} utilize convolutional neural network (CNN) to incorporate sparse LiDAR depth into the estimation from SGM~\cite{stereosgm} of stereo matching.
However, we argue that the sensor fusion directly applied to the depth outputs is not able to resolve the ambiguous correspondences existing in the procedure of stereo matching. Therefore, in this paper we advance to encode sparse LiDAR depth at earlier stages in stereo matching, i.e., matching cost computation and cost regularization, based on our proposed \textbf{\modelNameCBN}and \textbf{\modelNameIncat}techniques.}

\walon{
{\flushleft {\bf Conditional Batch Normalization.}} 
While the Batch Normalization layer improves network training via normalizing neural activations according to the statistics of each mini-batch, the Conditional Batch Normalization (CBN) operation instead learns to predict the normalization parameters (i.e., feature-wise affine transformation) in dependence on some conditional input. CBN has shown its generability in various application for coordinating different sources of information into joint learning. For instance, \cite{cbnreasoning} and \cite{cbnlanguage} utilize CBN to modulate imaging features by a linguistic embedding and successfully prove its efficacy for visual question answering. Perez \etal~\cite{cbnfilm} further generalize the CBN idea and point out its connections to other conditioning mechanisms, such as concatenation~\cite{conditiondcgan}, gating features~\cite{lstm}, and hypernetworks~\cite{hypernetworks}. \hubert{Lin \etal~\cite{lin2019cocogan} introduces CBN to a task of generating patches with spatial coordinates as conditions, which shares similar concept of modulating features by spatial-related information.} In our proposed method for the fusion of stereo camera and LiDAR sensor, we adopt the mechanism of CBN to integrate LiDAR data into the cost volume regularization step of the stereo matching framework, not only because of its effectiveness but also the clear motivation on reducing the search space of matching for more reliable disparity estimation.
}

\section{Method}

\begin{figure*}[t!]
\centering
\includegraphics[width=1.0\textwidth]{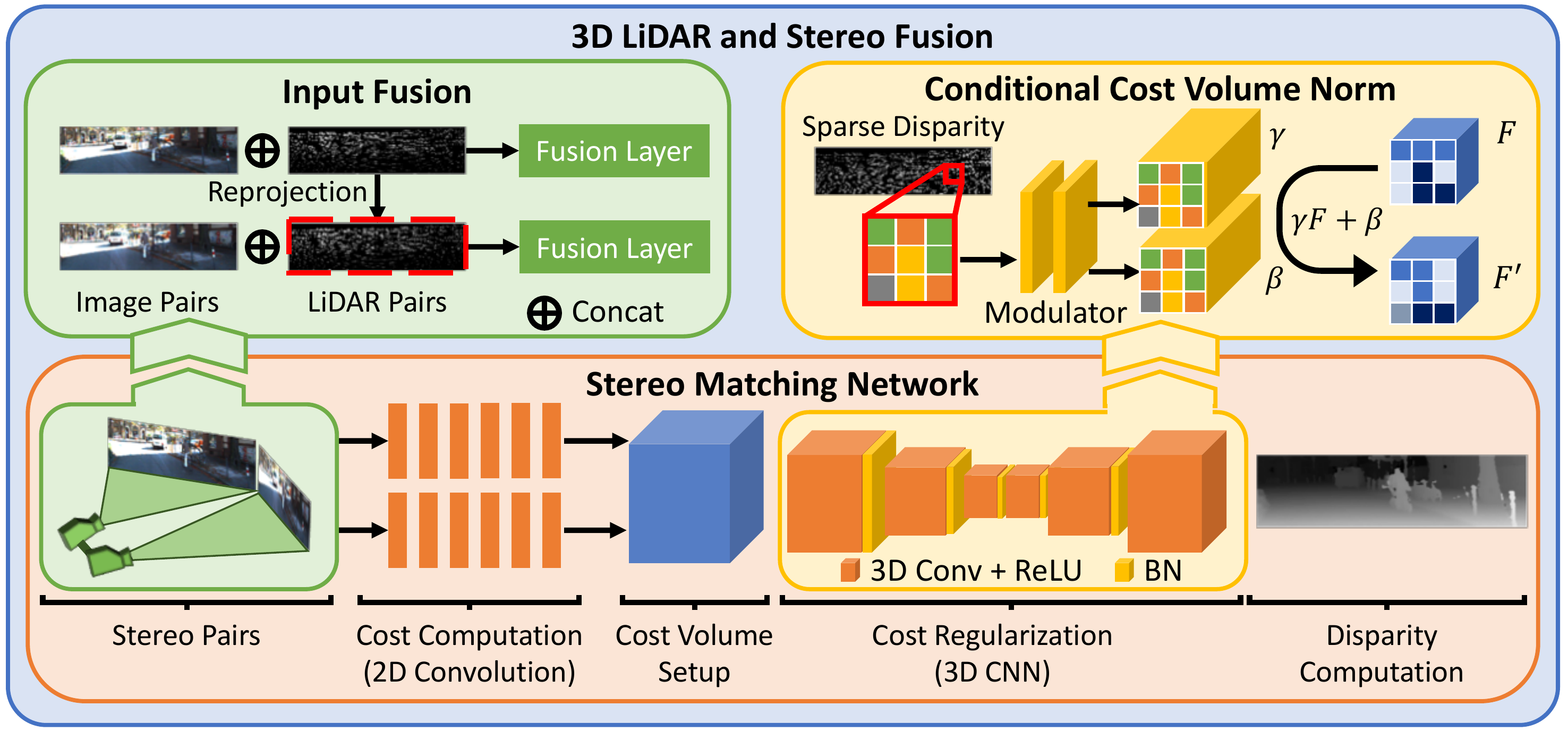}
\caption{\small{
Overview of our 3D LiDAR and stereo fusion framework. We introduce (1) \textbf{\modelNameIncatPunc} that incorporates the geometric information from sparse LiDAR depth with the RGB images as the input for the Cost Computation phase to learn joint feature representations, and (2) \textbf{\modelNameCBNPunc} that replaces batch normalization (BN) layer and modulates the cost volume features $F$ with being conditioned on LiDAR data, in the Cost Regularization phase of stereo matching network. With the proposed two techniques, Disparity Computation phase yields disparity estimate of high-precision.
%Overview of our proposed 3D LiDAR and stereo fusion pipeline.
%We introduce two mechanisms, \modelNameIncat and \modelNameCBNPunc, to incorporate sparse depth information with a stereo matching algorithm.
%In Stereo Input phase, we re-project a single LiDAR sweep to both left and right camera coordinates and triangulate them forming sparse disparity maps.
%\modelNameIncat leverages a fusion of context and geometry information by taking the concatenation of an RGB image and its re-projected LiDAR disparity map as input of Cost Computation phase.
%In Cost Regularization phase, we replace batch normalization (BN) layer with proposed \modelNameCBN layer.
%The modulator generates normalization parameters $\gamma$ and $\beta$ conditions on pixel values of a sparse disparity map. 
%\modelNameCBN modulates the cost volume features $F$ with parameters $\gamma$ and $\beta$.
%With the two techniques reasoning LiDAR and stereo fusion, Disparity Computation phase yields high-precision disparity maps.
}}
\label{fig:overview}
\end{figure*}

\walon{
As motivated above, we propose to fuse 3D LiDAR data into a stereo matching network by using two techniques: \modelNameIncat and \modelNameCBNPunc. In the following, we will first describe the baseline stereo matching network, and then sequentially provide the details of our proposed techniques. Finally, we introduce a hierarchical extension of \modelNameCBN which is more efficient in terms of runtime and memory consumption. The overview of our proposed method is illustrated in \figref{fig:overview}.

\subsection{Preliminaries of Stereo Matching Network}{
The end-to-end differentiable stereo matching network used in our proposed method, as shown in the bottom part of \figref{fig:overview}, is based on the work of GC-Net~\cite{stereogcnet} and is composed of four primary components which are in line with the typical pipeline of stereo matching algorithms~\cite{stereotaxonomy}. First, the deep feature extracted from a rectified left-right stereo pair is learned to compute the cost of stereo matching. The representation with encoded context information acts as a similarity measurement that is more robust than simple photometric appearance, and thus it benefits the estimation of pixel matches across stereo images. A cost volume is then constructed by aggregating the deep features extracted from the left-image with their corresponding ones from the right-image across each disparity level, where the size of cost volume is 4-dimensional $C\times H\times W\times D$ (i.e., \textit{feature~size $\times$ height $\times$ width $\times$ disparity}). To be detailed, the cost volume actually includes all the potential matches across stereo images and hence serves as a search space of matching. Afterwards, a sequence of 3D convolutional operations (3D-CNN) is applied for cost volume regularization and the final disparity estimation is carried out by regression with respect to the output volume of 3D-CNN along the $D$ dimension.
% constructing a cost volume involves aggregating left deep unary features with their corresponding right unary across each disparity level, forming a 4D volume ($feature~size\times height\times width \times disparity$). Subsequently, a sequence of 3D convolutional operations are carried out for cost volume regularization. Finally, we may estimate disparity from the final volume by reduction over the disparity dimension without post-processing. In this paper, we build our method upon GC-Net (Geometry and Context Network)~\cite{stereogcnet}.
}

\subsection{\modelNameIncat}{
\label{ssec:incat}
In the cost computation stage of stereo matching network, both left and right images of a stereo pair are passed through layers of convolutions for extracting features. In order to enrich the representation by jointly reasoning on appearance and geometry information from RGB images and LiDAR data respectively, we propose \modelNameIncat that simply concatenates stereo images with their corresponding sparse LiDAR depth maps. Different from~\cite{depths2d} that has explored a similar idea, for the setting of stereo and LiDAR fusion, we form the two sparse LiDAR depth maps corresponding to stereo images by reprojecting the LiDAR sweep to both left and right image coordinates with triangulation for converting depth values into disparity ones.
% In stereo matching networks, both left and right rectified stereo images are passed through a sequence of 2D convolutional operations to extract appearance information. In stereo and LiDAR fusion, we re-project a single LiDAR sweep to both left and right image coordinates to form two sparse depth maps, followed by triangulation that converts depth to disparity. Inspired by~\cite{depths2d}, we propose \modelNameIncat that simply concatenates left and right RGB images and their corresponding re-projected sparse disparity maps. By jointly reasoning on context and geometry information, the network can learn a more powerful feature representation for matching cost computation.
}

\subsection{Conditional Cost Volume Normalization (\modelNameCBNPunc)}{
\label{ssec:ccvnorm}
% We apply the aforementioned \modelNameIncat to generate deep unary features for matching cost computation by forming a cost volume. Subsequently, to better leverage LiDAR depth points, we suggest incorporating sparse depth information in cost regularization step, i.e. 3D-CNN in stereo matching network. 
In addition to \modelNameIncatPunc, we propose to incorporate information of sparse LiDAR depth points into the cost regularization step (i.e., 3D-CNN) of stereo matching network, learning to reduce the search space of matching and resolve ambiguities. As inspired by Conditional Batch Normalization (CBN)~\cite{cbnlanguage}\cite{cbnreasoning}, we propose \modelNameCBN (\textbf{Co}nditional \textbf{Co}st \textbf{Vo}lume \textbf{No}rmalization) to encode the sparse LiDAR information $L^{s}$ into the features of 4D cost volume $F$ of size $C\times H\times W\times D$. Given a mini-batch $\mathcal{B}=\{F_{i,\cdot,\cdot,\cdot,\cdot}\}_{i=1}^{N}$ composed of $N$ examples, 3D Batch Normalization (BN) is defined at training time as follows:
\begin{equation}
	\begin{split}
	F^{BN}_{i,c,h,w,d} &= \gamma_c \frac{F_{i,c,h,w,d}-\expectation_{\mathcal{B}}[F_{\cdot,c,\cdot,\cdot,\cdot}]}{\sqrt{Var_{\mathcal{B}}[F_{\cdot,c,\cdot,\cdot,\cdot}]+\epsilon}} + \beta_c
    \end{split}
    \label{eq:bn}
\end{equation}
\noindent where $\epsilon$ is a small constant for numerical stability and $\{\gamma_c, \beta_c\}$ are learnable BN parameters. 
When it comes to Conditional Batch Normalization, the new BN parameters $\{\gamma_{i,c}, \beta_{i,c}\}$ are defined as functions of conditional information $L^s_i$, for modulating the feature maps of cost volume in dependence on the given LiDAR data:
% On the other hand, CBN learns to generate new BN parameters $\gamma_{i,c}$ and $\beta_{i,c}$ as functions of conditional label $y_i$:
\begin{equation}
	\begin{split}
	\gamma_{i,c}=g_c(L^s_i),~~~~~~~\beta_{i,c}=h_c(L^s_i)
    \end{split}
    \label{eq:cbn}
\end{equation}
% \noindent  Conditioned on given inputs, CBN modulates the feature map by scaling, shifting, negating, or shutting them off, etc. 
However, directly applying typical CBN to 3D-CNN in stereo matching networks could be problematic due to few considerations: (1) Different from previous works~\cite{cbnlanguage}\cite{cbnreasoning}, the conditional input in our setting is a sparse map $L^s$ with varying values across pixels, which implies that normalization parameters should be carried out pixel-wisely; (2) An alternative strategy is required to tackle the void information contained in the sparse map $L^s$; (3) A valid value in $L^s_{h,w}$ should contribute differently to each disparity level of the cost volume.

Therefore, we introduce \modelNameCBNPunc ~(as shown in bottom-left of \figref{fig:normalization}) which better coordinates the 3D LiDAR information with the nature of cost volume to tackle the aforementioned issues: 
% which is in consonance with the nature of cost volume and its relation to additional disparity information:
\begin{equation}
    \small
	\begin{split}
	F^{CCVNorm}_{i,c,h,w,d} &= \gamma_{i,c,h,w,d} \frac{F_{i,c,h,w,d}-\expectation_{\mathcal{B}}[F_{\cdot,c,\cdot,\cdot,\cdot}]}{\sqrt{Var_{\mathcal{B}}[F_{\cdot,c,\cdot,\cdot,\cdot}]+\epsilon}} + \beta_{i,c,h,w,d} \\
	\gamma_{i,c,h,w,d} &=
	\begin{cases}
	g_{c,d}(L^s_{i,h,w}),& \text{if }L^s_{i,h,w}\text{ is valid}\\
	\overline{g}_{c,d},& \text{otherwise}
	\end{cases} \\
	\beta_{i,c,h,w,d} &=
	\begin{cases}
	h_{c,d}(L^s_{i,h,w}),& \text{if }L^s_{i,h,w}\text{ is valid}\\
	\overline{h}_{c,d},& \text{otherwise}
	\end{cases}
    \end{split}
    \label{eq:cocovono}
\end{equation}
\noindent Intuitively, given a LiDAR point $L^s_{h,w}$ with a valid value, the representation (i.e., $F_{c,h,w,d}$) of its corresponding pixel in the cost volume under a certain disparity level $d$ would be enhanced/suppressed via the conditional modulation when the depth value of $L^s_{h,w}$ is consistent/inconsistent with $d$. In contrast, for those LiDAR points with invalid values, the regularization upon the cost volume degenerates back to a unconditional batch normalization version and the same modulation parameters $\{\overline{g}_{c,d}, \overline{h}_{c,d}\}$ are applied to them. 
We experiment the following two different choices for modelling the functions $g_{c,d}$ and $h_{c,d}$:

{\noindent {\bf Categorical \modelNameCBNPunc}}: a $\hat{D}$-entry lookup table with each element as a $D\times C$ vector is constructed to map LiDAR values into normalization parameters $\{\gamma, \beta\}$ of different feature channels and disparity levels, where the LiDAR depth values are discretized here into $\hat{D}$ levels as entry indexes. 
% selected by a simple indexing operation based on the discrete disparity, which is obtained by discretizing sparse disparity into $\hat{D}$ levels.

{\noindent {\bf Continuous \modelNameCBNPunc}}: a CNN is utilized to model the continuous mapping between the sparse LiDAR data $L^{s}$ and the normalization parameters of $D\times C$-channels. In our implementation, we use the first block of ResNet34~\cite{resnet} to encode LiDAR data, followed by one $1\times 1$ convolution for \modelNameCBN in different layers respectively. 

%In Eq.\eqref{eq:cocovono}, we propose to experiment two different choices for modelling the conditioning functions $g_{c,d}$ and $h_{c,d}$. First, we can formulate them as a $D$-entry lookup table with each element as a $D\times C$ vector selected by a simple indexing operation based on discretized disparity, denoted as Categorical \modelNameCBNPunc. Second, we can utilize a CNN to model the continuous mapping between disparity and the modulation parameters, denoted as Continuous \modelNameCBNPunc. 
}

\subsection{Hierarchical Extension}{
% the $M$-layer continuous variants needs $\Sigma_{j=1}^{M-1}k_{j}k_{j+1}m_{j}m_{j+1}+2DC$ parameters, where $k$ is kernel size, $m$ is number of channels, $m_1=1$ and $m_M=DC$.
We observe that both Categorical and Continuous \modelNameCBNPunc~require a huge number of parameters. For each normalization layer, the Categorical version demands $\mathcal{O}(\hat{D}DC)$ parameters to build up the lookup table while the CNN for Continuous one even needs more for desirable performance. In order to reduce the model size for practical usage, we advance to propose a hierarchical extension (denoted as \textbf{\modelNameHierCBNPunc}, which is shown in the top-right of \figref{fig:normalization}), serving as an approximation of the Categorical \modelNameCBN with much fewer model parameters. The normalization parameters of \modelNameHierCBN for valid LiDAR points are computed by:
\begin{equation}
	\begin{split}
	\gamma_{i,c,h,w,d} &= \phi^g(d)g_c(L^s_{i,h,w})+\psi^g(d) \\
	\beta_{i,c,h,w,d} &= \phi^h(d)h_c(L^s_{i,h,w})+\psi^h(d)
    \end{split}
    \label{eq:hier_cocovono}
\end{equation}
% \noindent The basic concept is to modulate the lower-dimensional normalization parameters $g_c(\cdot)$ 
% %($DC$ for categorical version and $\Sigma_{i=1}^{N-2}m_{i}m_{i+1}+m_{N-1}C$ for continuous version)
% by another pair of modulation parameters $\phi^g(d)$ and $\psi^g(d)$. Note that $\phi^g, \psi^g, \phi^h, \psi^h$ are, by their nature, lookup tables (with the size of $D\times C$) since their inputs are now constrained to the number of disparity levels of the cost volume. With this hierarchical approximation, for each normalization layer it requires only $\mathcal{O}(DC)$ parameters.
\noindent Basically, the procedure of mapping from LiDAR disparity to a $D\times C$ vector in Categorical \modelNameCBN is now decomposed into two sequential steps. Take $\gamma$ for an example, $g_c$ is first used to compute the intermediate representation (i.e., a vector in size $C$) conditioned on $L_{i,h,w}^s$, and is then modulated by another pair of modulation parameters $\{\phi^g(d), \psi^g(d)\}$ to obtain the final normalization parameter $\gamma$. Note that $\phi^g, \psi^g, \phi^h, \psi^h$ are basically the lookup table with the size of $D\times C$. With this hierarchical approximation, each normalization layer only requires $\mathcal{O}(DC)$ parameters.

% After modification, for each layer, hierarchical Categorical \modelNameCBN requires $2\hat{D}C+6DC$ parameters, while the continuous version needs $\Sigma_{j=1}^{M-2}k_{j}k_{j+1}m_{j}m_{j+1}+m_{M-1}C+6DC$ parameters.
}
}

\begin{figure}[t]
\centering
\includegraphics[width=1.0\linewidth]{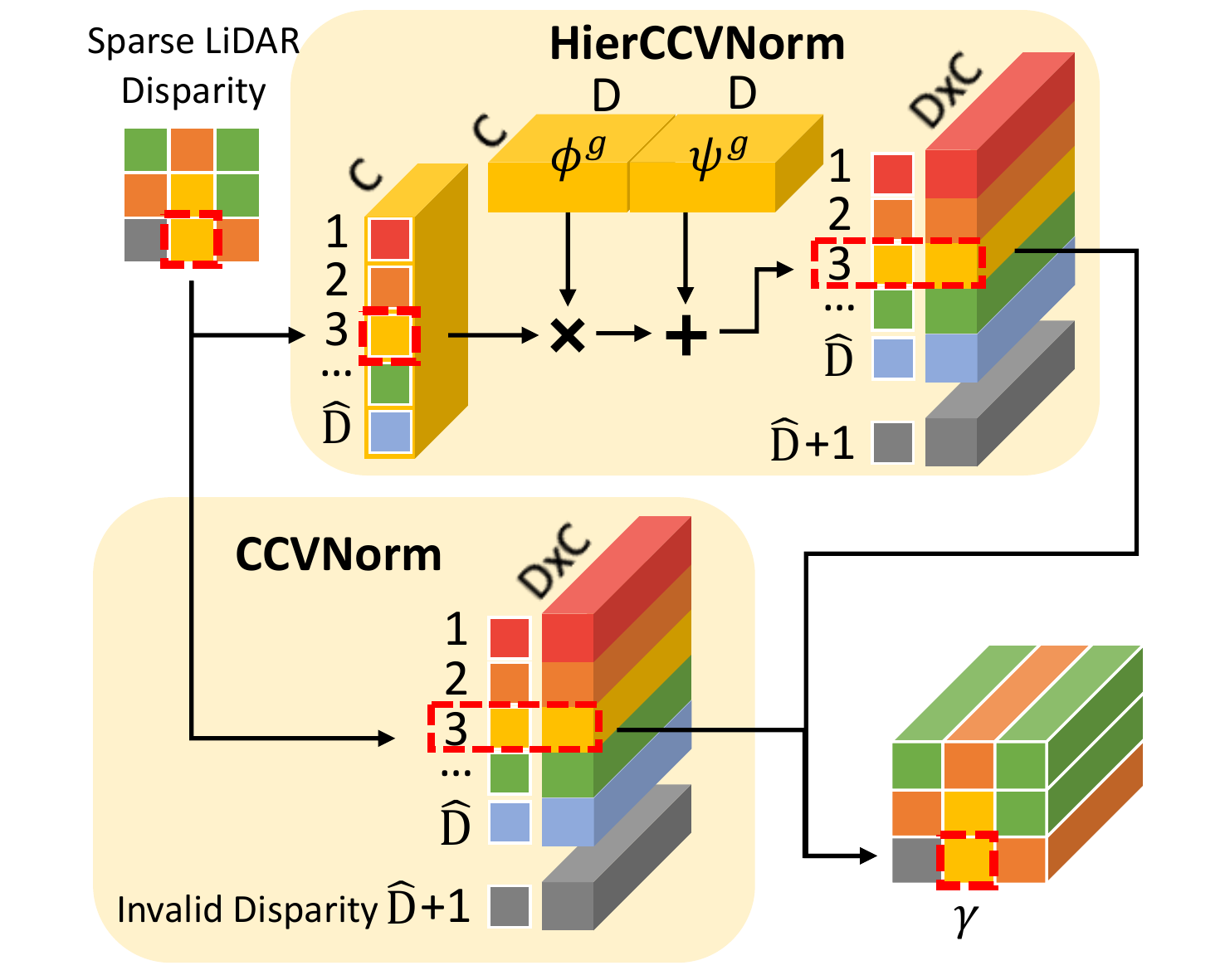}
\caption{\small{
% Conditional Cost Volume Normalization. At each pixel (red dashed bounding box), categorical \modelNameCBN selects from a $\hat{D}$-entry lookup table to form the modulation parameters $\gamma$ based on the discretized disparity value, with invalid points are separately handled with an additional set of parameters (gray). Besides, \modelNameHierCBN produces $\gamma$ by hierarchically modulating the condition $g(\cdot)$ with another set of modulation parameters $\{\gamma^g, \beta^g\}$.
Conditional Cost Volume Normalization. At each pixel (red dashed bounding box), based on the discretized disparity value of corresponding LiDAR data, categorical \modelNameCBN selects the modulation parameters $\gamma$ from a $\hat{D}$-entry lookup table, while the LiDAR points with invalid value are separately handled with an additional set of parameters (in gray color). On the other hand, \modelNameHierCBN produces $\gamma$ by a hierarchical modulation of 2 steps, with modulation parameters $g_c(\cdot)$ and $\{\phi^g, \psi^g\}$ respectively (cf. Eq.~\ref{eq:hier_cocovono}).
%Conditional Cost Volume Normalization. 
%Given the sparse disparity pattern, our proposed \modelNameCBN modulates cost volume with normalization parameters $\gamma$ and $\beta$ conditioning on the disparity value pixel-wisely.
%In addition to $D$ disparity level, we regulate the invalid disparity points separately by introducing an additional pair of parameters.
%\modelNameCBN selects $D \times C$ parameters based on discretized disparity values while \modelNameHierCBN picks only $C$ parameters which modulated by another parameter pair $\gamma^g~and~\beta^g$. $g_c(\cdot)$
}}
\label{fig:normalization}
\end{figure}

\section{Experimental Results}

% \johnson{
% In this section, we first compare our method to other baselines on two KITTI dataset~\cite{kitti2015}\cite{kitti2017}, coming up with \hubert{a series of} ablation studies to step-by-step demonstrate the effectiveness of the proposed method. 
% Next, we examine the robustness of our method to LiDAR density and \hubert{analyze the qualitative results against multiple baselines}. 
% Finally, we benchmark the computation time of our approach.
% }

\walon{
We evaluate the proposed method on two KITTI datasets~\cite{kitti2015}\cite{kitti2017} and show that our framework is able to achieve favorable performance in comparison with several baselines. In addition, we extensively conduct a series of ablation study to sequentially demonstrate the effectiveness of our design choices in the proposed method. Moreover, we investigate the robustness of our approach with respect to the density of LiDAR data, as well as benchmark the runtime and memory consumption. The code and model will be made available for the public.}

\subsection{Experimental Settings}{
\walon{
{\flushleft {\bf KITTI Stereo 2015 Dataset.} }
KITTI Stereo dataset~\cite{kitti2015} is commonly used for evaluating stereo matching algorithms. It contains $200$ stereo pairs for each of training and testing set, where the images are in size of $1242\times 375$. As the ground truth is only provided for the training set, we follow the identical setting as previous works~\cite{LiDARstereoprobfusion}\cite{LiDARstereoicra18} to evaluate our model on the training set with LiDAR data. For model training, since only $142$ pairs among the training set are associated with LiDAR scans and they cover $29$ scenes in the KITTI Completion dataset~\cite{kitti2017}, we hence train our network on the subset of the Completion dataset with images of non-overlapping scenes (i.e., $33$k image pairs remained for training).
% where the images in the overlapped 29 scenes are filtered out and the remaining dataset contains 33k image pairs.

{\flushleft {\bf KITTI Depth Completion Dataset.}}
KITTI Depth Completion dataset~\cite{kitti2017} collects semi-dense ground truth of LiDAR depth map by aggregating $11$ consecutive LiDAR sweeps together, with roughly $30\%$ pixels annotated. The dataset consists of $43$k image pairs for training, $3$k for validation, and $1$k for testing. Since no ground truth is available in the testing set, we split the validation set into $1$k pairs for validation and another $1$k pairs for testing that contain non-overlapped scenes with respect to the training set.
% As a result, our testing set is different from the one provided by the dataset.
We also note that the full-resolution images (in size of $1216\times 352$) of this dataset are bottom-cropped to $1216\times 256$ because there is no ground truth on the top.

% KITTI Depth Completion dataset~\cite{kitti2017} is a commonly used benchmark in depth completion~\cite{depths2d}\cite{depthrgbguide}\cite{depthpnp}. 
% The dataset is created by aggregating 11 consecutive LiDAR sweeps into one, generating a semi-dense ground truth with roughly 30\% annotated pixels.
% The full-resolution images ($1216\times 352$) are bottom-cropped to $1216\times 256$ because there is no ground truth in the top. 
% The dataset consists of 43k image pairs training data, 3k for validation data, and 1,000 testing data without ground truth. 
% Since the \hubert{original} benchmark is for monocular depth estimation and we have no access to \hubert{the ground truth of testing data to evaluate models in our setting}, we split the 3k validation set to 1,000 pairs for our validation and another 1,000 pairs from non-overlapped scenes for testing. 
% \hubert{As a result,} our testing set is different from the testing set provided by the dataset.

{\flushleft {\bf Evaluation Metric.} }
We adopt standard metrics in stereo matching and depth estimation respectively for the two datasets: (1) On KITTI Stereo~\cite{kitti2015}, we follow its development kit to compute the percentage of disparity error that is greater than 1, 2 and 3 pixel(s) away from the ground truth; (2) On KITTI Depth Completion~\cite{kitti2017}, Root Mean Square Error (RMSE), Mean Absolute Error (MAE), and their inverse ones (i.e., iRMSE and iMAE) are used. 

{\flushleft {\bf Implementation Details.}}
Our implementation is based on PyTorch and follows the training setting of GC-Net~\cite{stereogcnet} to have $\mathcal{L}1$ loss for disparity estimation. The optimizer is RMSProp~\cite{rmsprop} with a constant learning rate $1\times 10^{-3}$. The model is trained with batch size of $1$ using a randomly-cropped $512\times 256$ image for $170$k iterations.
% and we validate our model using checkpoints in the last $20$k iterations.
The maximum disparity is set to $192$. We apply \modelNameCBN to the 21, 24, 27, 30, 33, 34, 35th layers in GC-Net. We note that our full model refers to the setting of having both \modelNameIncat and \modelNameHierCBNPunc, unless otherwise specified.
}

\subsection{Evaluation on the KITTI Datasets}{

\begin{table}[!t]
\centering
\caption{\small{Evaluation on the KITTI Stereo 2015 Dataset.}}
\begin{tabular}{l l | c c c c}
\hline
Method & Sparsity & $>$ 3 px~$\shortdownarrow$ & $>$ 2 px~$\shortdownarrow$ & $>$ 1 px~$\shortdownarrow$ \\ \hline \hline
SGM~\cite{stereosgm} & \multirow{3}{*}{None} & $20.7$ & - & - \\
MC-CNN~\cite{stereomccnn} & & $6.34$ & - & - \\ 
GC-Net~\cite{stereogcnet} & & $4.24$ & $5.82$ & $9.97$ \\ \hline
Prob. Fusion~\cite{LiDARstereoprobfusion} & \multirow{3}{*}{\makecell[l]{LiDAR\\ Data}} & $5.91$ & - & - \\
Park \textit{et al.}~\cite{LiDARstereoicra18} & & $4.84$ & - & -\\
Ours Full & & $\mathbf{3.35}$ & $\mathbf{4.38}$ & $\mathbf{6.79}$ \\
\hline
\end{tabular}
\small
\label{tab:kitti_stereo}
\end{table}

%\footnotemark[\value{footnote}]
%\footnotetext{Our final model is \modelNameIncat + Categorical \modelNameHierCBNPunc.}

\walon{
For the KITTI Stereo 2015 dataset, we compare our proposed method to several baselines of stereo matching and LiDAR fusion in \tabref{tab:kitti_stereo}. We draw few observations here: $1)$ Without using any LiDAR data, deep learning-based stereo matching algorithms (i.e., MC-CNN~\cite{stereomccnn} and GC-Net~\cite{stereogcnet}) perform better than the conventional one (i.e., SGM~\cite{stereosgm}) by a large margin; $2)$ GC-Net outperforms MC-CNN since its entire stereo matching process is formulated in an end-to-end learning framework, and it even performs competitively compared to two other baselines having LiDAR data fused either in input or output spaces (i.e., Probabilistic Fusion~\cite{LiDARstereoprobfusion} and Park \etal~\cite{LiDARstereoicra18} respectively). This observation shows the importance of using an end-to-end trainable stereo matching network as well as designing a proper fusion scheme; $3)$ Our full model learns to well leverage the LiDAR information into both the matching cost computation and cost regularization stages of the stereo matching network and obtains the best accuracy for disparity estimation against all the baselines.

In addition to disparity estimation, we compare our model with both monocular depth completion approaches and fusion methods of stereo and LiDAR data on the KITTI Completion dataset in \tabref{tab:kitti_completion}. From the results of Park~\etal, we observe that even with more information from stereo pairs, the performance is not guaranteed to be better than state-of-the-art method for monocular depth completion (i.e., NConv-CNN~\cite{depthnconv}, Ma \etal~\cite{depthsss2d}, and FusionNet~\cite{depthrgbguide}) if the stereo images and LiDAR data are not properly integrated. On the contrary, our method with careful designs of the proposed \modelNameIncat and \modelNameHierCBNPunc~is able to outperform baselines of both monocular or stereo fusion. It is also worth noting that, our model shows significant boost on the metrics related to inverse depth (i.e., iRMSE and iMAE) since our method is trained to predict disparity. Particularly, we emphasize here the importance of the inverse depth metrics, since they demand higher accuracy in the closer region, which are especially suitable for robotic tasks.

% The first thing to be remarked is that distinct from depth completion scenario, we are provided with an extra right RGB image in stereo and LiDAR fusion. 
% Though this inherently implies that stereo and LiDAR fusion methods should bring better performance, it is still worthy to demonstrate the comparison as these two settings are both widely taken on in robotics applications~\cite{LiDARstereomobile}\cite{LiDARstereorecon}. 
% The monocular depth completion methods \hubert{that compared with ours} are all top-ranked approaches in the KITTI Completion benchmark by the time of this paper is written. 

% Our method greatly exceeds other monocular baselines in iRMSE and iMAE. 
% Kindly note that our stereo matching network is trained on predicting disparity, which is the inverse of depth and, in such case, low disparity error in a pixel may correspond to both low or high depth error depending on the depth range the pixel locates in. 
% Accordingly, our method only performs slightly better than others in RMSE and is worse in MAE. 
% Furthermore, we implement Park \textit{et al.}~\cite{LiDARstereoicra18} on KITTI Completion Dataset. 
% For disparity metrics, they achieve 1.17 in 3-pixel error. 
% However, they perform badly in depth metrics and are outperformed by our method. 
% Lastly, we emphasize the importance of inverse depth metrics, since they demand higher accuracy in the closer region, which are especially suitable error measures for robotics tasks.
}
}

\begin{table}[!t]
\centering\scriptsize
\caption{\small{Evaluation on the KITTI Depth Completion Dataset.}}
\begin{tabular}{l l | c c c c}
\hline
Data & Method & iRMSE~$\shortdownarrow$ & iMAE~$\shortdownarrow$ & RMSE~$\shortdownarrow$ & MAE~$\shortdownarrow$ \\ \hline \hline
\multirow{3}{*}{Mono} & NConv-CNN~\cite{depthnconv} & $2.60$ & $1.03$ & $0.8299$ & $0.2333$ \\
 & Ma \textit{et al.}~\cite{depthsss2d} & $2.80$ & $1.21$ & $0.8147$ & $0.2499$ \\
 & FusionNet~\cite{depthrgbguide} & $2.19$ & $0.93$ & $0.7728$ & $\mathbf{0.2150}$ \\ \hline
\multirow{2}{*}{Stereo} & Park \textit{et al.}~\cite{LiDARstereoicra18} & $3.39$ & $1.38$ & $2.0212$ & $0.5005$ \\
% & Ours (Non-hier.) & $\mathbf{1.3940}$ & $\mathbf{0.8049}$ & $\mathbf{0.7325}$ & $0.2501$ \\
% & Ours & $\mathbf{1.3968}$ & $\mathbf{0.8069}$ & $\mathbf{0.7493}$ & $0.2525$ \\
 & Ours Full & $\mathbf{1.40}$ & $\mathbf{0.81}$ & $\mathbf{0.7493}$ & $0.2525$ \\
\hline
\end{tabular}
\small
\label{tab:kitti_completion}
\end{table}

\subsection{Ablation Study}{

\begin{table*}[t]
\centering
\caption{\small{Ablation study on the KITTI Depth Completion Dataset. ``IF'', ``Cat'', and ``Cont'' stand for \textbf{I}nput \textbf{F}usion, categorical and continuous variants of \modelNameCBNPunc, respectively. For different stages, ``MCC'' stands for \textbf{M}atching \textbf{C}ost \textbf{C}omputation and ``CR'' is \textbf{C}ost \textbf{R}egularization. The bold font indicates top-2 performance.
% \hubert{Note that ``IF`` stands for ``\textbf{I}nput \textbf{F}usion``, ``MCC'' stands for ``\textbf{M}atching \textbf{C}ost \textbf{C}omputation'' stage and ``CR'' is ``\textbf{C}ost \textbf{R}egularization'' stage}.
}
}
\begin{tabular}{l | c | c c c c c | c c c c}
\hline
\multirow{2}{*}{Method} & \multirow{2}{*}{Stage} & \multicolumn{5}{c}{Disparity} & \multicolumn{4}{|c}{Depth} \\
 & & $>$ 3 px~$\shortdownarrow$ & $>$ 2 px~$\shortdownarrow$ & $>$ 1 px~$\shortdownarrow$ & RMSE~$\shortdownarrow$ & MAE~$\shortdownarrow$ & RMSE~$\shortdownarrow$ & MAE~$\shortdownarrow$ & iRMSE~$\shortdownarrow$ & iMAE~$\shortdownarrow$ \\ \hline \hline
GC-Net~\cite{stereogcnet} & \multirow{2}{*}{MCC} & $0.2540$ & $0.4303$ & $1.5024$ & $0.6526$ & $0.4020$ & $1.0314$ & $0.4054$ & $1.6814$ & $1.0356$ \\
+ IF &  & $0.1694$ & $0.3086$ & $1.0405$ & $0.5559$ & $0.3245$ & $0.7659$ & $0.2613$ & $1.4324$ & $0.8362$  \\ \hline
+ FeatureConcat & \multirow{5}{*}{CR} & $0.1810$ & $0.3227$ & $1.1335$ & $0.5946$ & $0.3812$ & $0.8791$ & $0.3443$ & $1.5318$ & $0.9821$  \\
+ NaiveCBN & & $0.2446$ & $0.4342$ & $1.5616$ & $0.6405$ & $0.3915$ & $1.0067$ & $0.3808$ & $1.6505$ & $1.0087$  \\
+ \modelNameCBN (Cont) &  & $0.1363$ & $0.2532$ & $1.0265$ & $0.5856$ & $0.3688$ & $0.8661$ & $0.3385$ & $1.5087$ & $0.9500$ \\
+ \modelNameCBN (Cat) &  & $0.1254$ & $0.2596$ & $1.1348$ & $0.5625$ & $0.3574$ & $0.8942$ & $0.3425$ & $1.4493$ & $0.9209$ \\
+ Hier\modelNameCBN (Cat) &  & $0.1268$ & $0.2592$ & $1.1124$ & $0.5615$ & $0.3583$ & $0.8898$ & $0.3403$ & $1.4466$ & $0.9230$  \\ \hline
+ IF + FeatureConcat & \multirow{4}{*}{\makecell{MCC\\+\\CR}} & $0.1578$ & $0.2958$ & $1.0012$ & $0.5550$ & $0.3256$ & $0.7622$ & $0.2643$ & $1.4303$ & $0.8389$ \\
+ IF + \modelNameCBN (Cont) &  & $0.1460$ & $0.2657$ & $0.9586$ & $0.6137$ & $0.3235$ & $0.7727$ & $0.2573$ & $1.5795$ & $0.8335$  \\
+ IF + \modelNameCBN (Cat) &  & $\mathbf{0.1194}$ & $\mathbf{0.2406}$ & $\mathbf{0.9227}$ & $\mathbf{0.5409}$ & $\mathbf{0.3124}$ & $\mathbf{0.7325}$ & $\mathbf{0.2501}$ & $\mathbf{1.3940}$ & $\mathbf{0.8049}$ \\
+ IF + Hier\modelNameCBN (Cat) &  & $\mathbf{0.1196}$ & $\mathbf{0.2457}$ & $\mathbf{0.9554}$ & $\mathbf{0.5420}$ & $\mathbf{0.3131}$ & $\mathbf{0.7493}$ & $\mathbf{0.2525}$ & $\mathbf{1.3968}$ & $\mathbf{0.8069}$  \\
\hline
\end{tabular}
\vspace{-1em}
\small
\label{tab:ablation}
\end{table*}
}

\johnson{
% [Comment] describe experiments and baselines
In \tabref{tab:ablation}, we show the effectiveness of the proposed components step-by-step. Two additional baselines for fusion are introduced to have more throughout comparison: \textbf{Feature Concat} and \textbf{Naive CBN}.
Feature Concat uses a ResNet34~\cite{resnet} to encode LiDAR data, as utilized in other depth completion methods~\cite{depths2d}\cite{depthsss2d}, and concatenate the LiDAR feature to the cost volume feature. 
Naive CBN follows a straightforward design of CBN that modulates the cost volume feature conditioned on valid LiDAR depth values. 
% For the continuous version of \modelNameCBNPunc, we use the first block of ResNet34 to encode LiDAR data, followed by one $1\times 1$ convolution for \modelNameCBN in different layers respectively. 

% [Comment] input fusion is good
{\flushleft {\bf Overall Results.}}
First, we find that \modelNameIncat significantly improves the performance comparing to GC-Net. 
This highlights the significance of incorporating geometry information in the early matching cost computation (MCC) stage, mentioned in \secref{ssec:incat}. 
% [Comment] compare baseline in cost regularization stage: feature concat, naive cbn, our variants
Next, in the cost regularization (CR) stage, we compare Feature Concat, Naive CBN, and different variants of our methods. 
All our \modelNameCBN variants outperform other mechanisms in fusing the LiDAR information to the cost volume in stereo matching networks.
This demonstrates the benefit of applying the proposed \modelNameCBN scheme which serves as a regularization step on feature fusion for facilitating stereo matching (\secref{ssec:ccvnorm}).
%
% [Comment] clarify what our final model is
Finally, our full models with \modelNameIncat and categorical \modelNameCBN (with and without the hierarchical extension) produce the best results in the ablation.
% affine transformation over a feature and the special design of \modelNameCBN tailored for stereo matching networks (\secref{ssec:ccvnorm}).

\begin{figure}[t!]
\centering
\includegraphics[width=\linewidth]{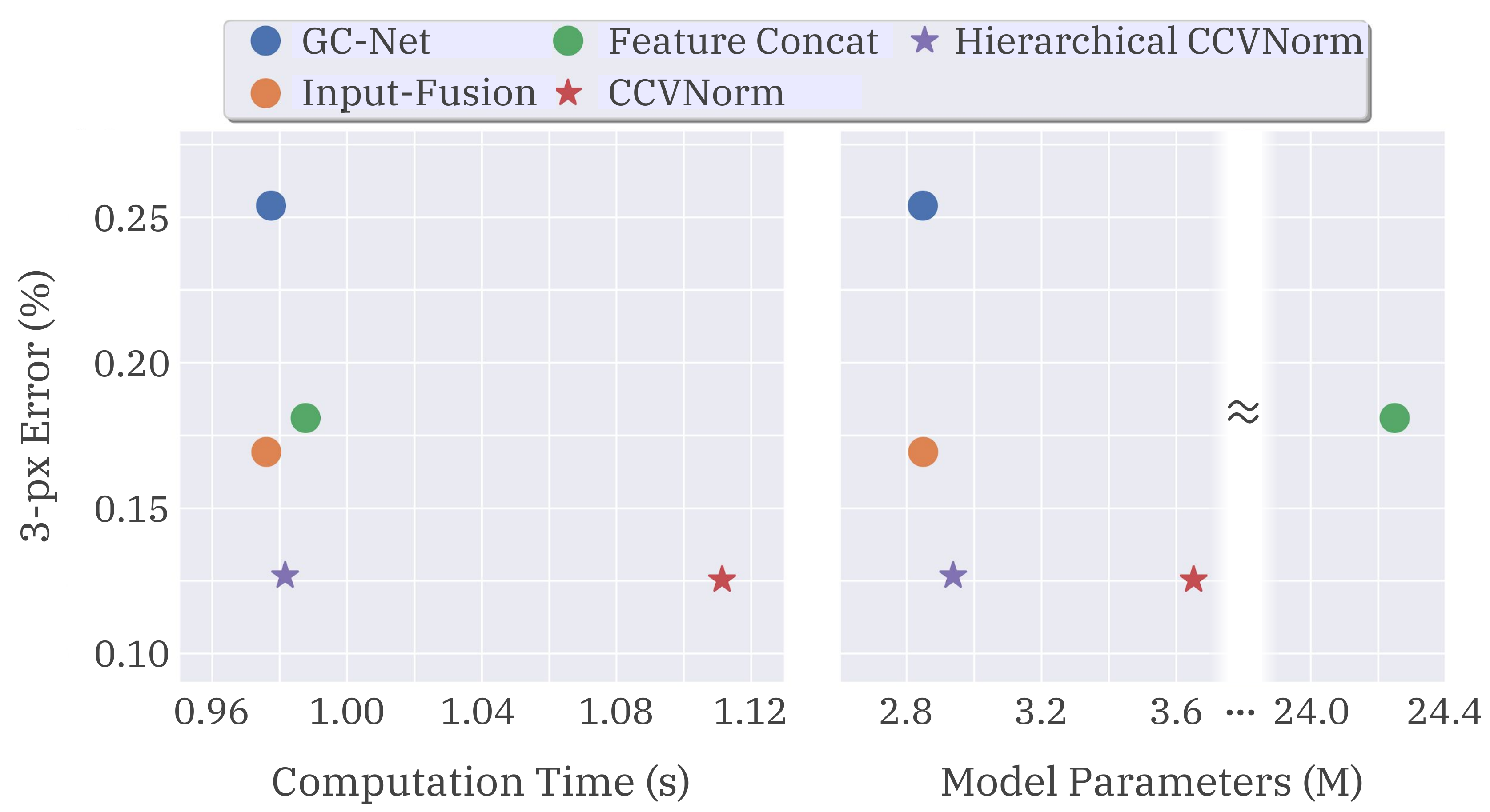}
\caption{\small{Error v.s. computation time and model parameters. It demonstrates that our hierarchical \modelNameCBN achieves comparable performance to the original \modelNameCBN but with much less overhead in computational time and model parameters.}}
\label{fig:hier-good}
\end{figure}

{\flushleft {\bf Categorical v.s. Continuous.}}
% [Comment] categorical better than continuous in most metrics
In addition, we empirically find that the categorical \modelNameCBN may serve as a better conditioning strategy than the continuous variant.
% [Comment] explain larger 1-px error in categorical version comparing to continuous version
Another interesting discovery is that the categorical variant performs competitively compared to the continuous one in most metrics (for disparity) except for the 1-px error. %This is not a unique phenomenon throughout all our experiments.
This is not surprising since the conditioning label for categorical \modelNameCBN is actually discretized LiDAR data, which may possibly lead to the propagation of quantization error. 
% [Comment] explain larger disparity RMSE and MAE in continuous version comparing to categorical version
While the continuous variant performs better in 1-px error, they may not necessarily yield better results in sub-pixel errors (i.e., disparity RMSE and MAE), since cost volume is naturally a discretization in the disparity space, thus making the continuous variant harder to handle sub-pixel predictions~\cite{stereogcnet}.
%While the continuous variant performs better in 1-px error, they may not necessarily yield results with better disparity RMSE and MAE, since the nature of cost volume makes it hard to handle sub-pixel error~\cite{stereogcnet}.%, which is commonly alleviated by specially-designed disparity computation techniques~\cite{stereogcnet}. %This suggests that the continuous variant may not necessarily yield results with better disparity RMSE and MAE.

{\flushleft {\bf Benefits of Hierarchical \modelNameCBNPunc.}}
% [Comment] our hierarchical extension is good
In \tabref{tab:ablation}, our hierarchical extension approximates the original \modelNameCBNPunc ~and achieves comparable performance.
% Hence, empirically hierarchical \modelNameCBN is capable of serving as an approximation of the original \modelNameCBNPunc. 
We further show the computational time and model size for various conditioning mechanisms in \figref{fig:hier-good} that highlights the advantages of our hierarchical \modelNameCBNPunc. 
In both sub-figures, the scattered point closer to the left-bottom corner indicates a more accurate and efficient model.
The figure shows that our hierarchical \modelNameCBN achieves good performance boost with only a small overhead in both computational time and model parameters compared to GC-Net. 
Note that, \textit{Feature Concat} adopts a standard strategy to encode LiDAR data in depth completion methods~\cite{depths2d}\cite{depthsss2d}, resulting in much more parameters introduced. 
Overall, our hierarchical extension can be viewed as an approximation of the original \modelNameCBN with a huge reduction of computational time and parameters.
}
}

\subsection{Robustness to LiDAR Density}{

\begin{figure}[t]
    \centering
    \includegraphics[width=\linewidth]{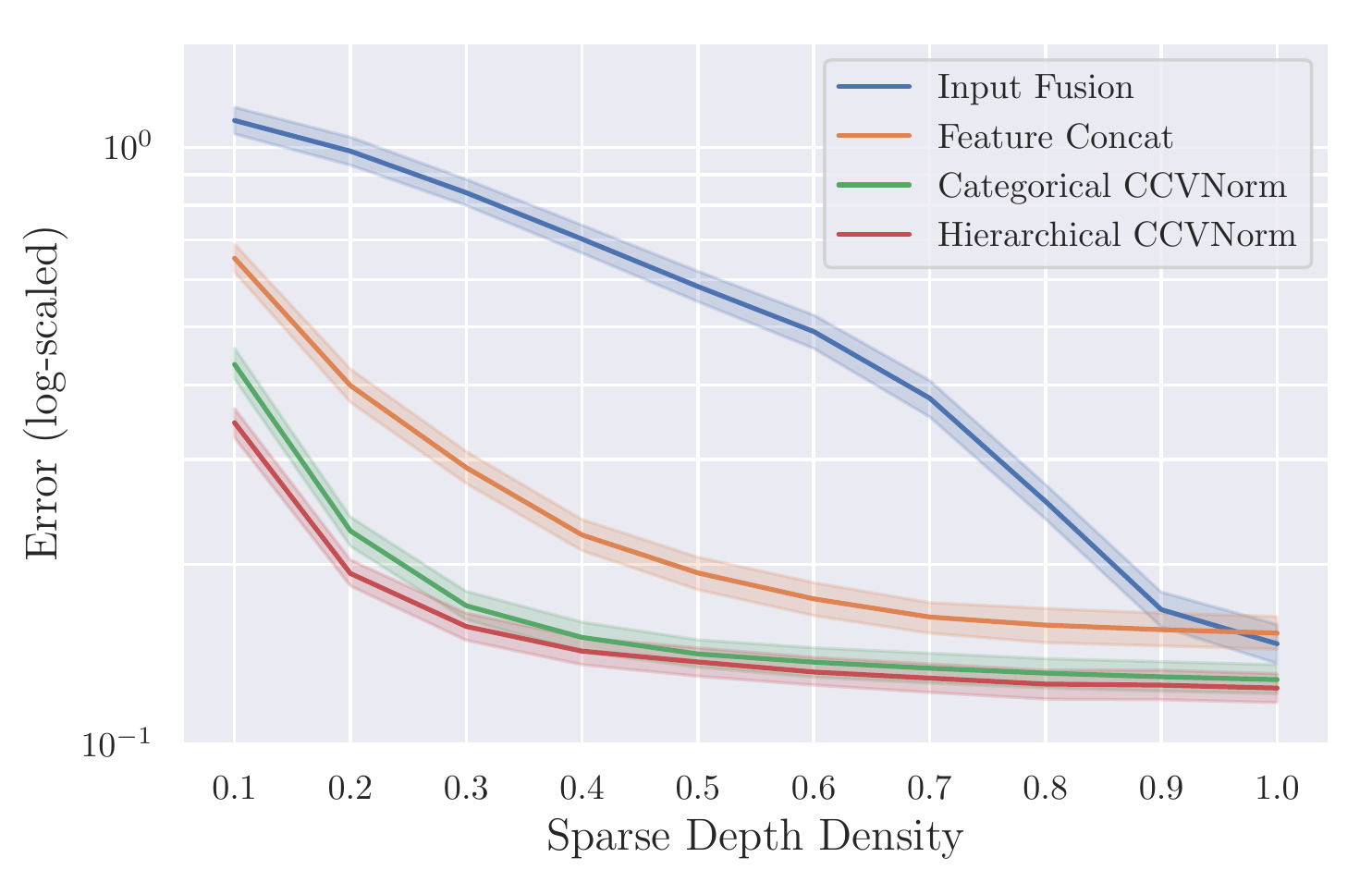}
    \caption{\small{Robustness to LiDAR density. The 1.0 value in the horizontal axis indicates a complete LiDAR sweep and the shadow indicates the standard deviation. The figure shows that our method is more robust to LiDAR sub-sampling comparing to other baselines.}}
    \label{fig:vis-robustness}
\end{figure}

\walon{
In \figref{fig:vis-robustness}, we study the robustness of different fusion mechanisms to the change of density in LiDAR data. We use $1.0$ on the horizontal axis of \figref{fig:vis-robustness} to indicate a full LiDAR sweep, and gradually sub-sample from it and observe how the performance of each fusion approach varies. The results highlight that both variants of \modelNameCBN (i.e., Categorical and Hierarchical \modelNameCBNPunc) are consistently more robust to different density levels in comparison with other baselines (i.e., Feature Concat and Input Fusion only).

% In \figref{fig:vis-robustness}, we investigate the robustness of different conditioning mechanisms to the change of LiDAR data density. 
% The rightmost label density $1.0$ in the horizontal axis indicates a full LiDAR sweep. 
% We gradually sub-sample from a complete LiDAR sweep and observe the performance drop. 
% \hubert{The results highlight that both variants of \modelNameCBN are consistently more robust to different density levels comparing to other baselines.} 
First, \modelNameIncat is highly sensitive to the density of sparse depth due to its property of treating both valid/invalid pixels equally and setting invalid values as a fixed constant, and hence introducing numerical instability during network training/inference.
% Since \modelNameIncat sets missing values to a fixed constant and treats both valid and invalid pixels equally, this introduces numerical instability during network inference and results in large performance drop after LiDAR sub-sampling. 
% Some related literature~\cite{kitti2017}\cite{depthhmsnet} propose special operations to specifically handle the sparse data. 
% Tackling such issue may potentially provide better performance, but we leave it to future work as it is not the main focus of this paper. 
Second, by comparing our two variants with Feature Concat, we observe that both methods do not suffer from severe performance drop in high LiDAR density ($0.7 \sim 1.0$). 
However, in low-level density ($0.1 \sim 0.6$), Feature Concat drastically degrades the performance while ours remains robust to the sub-sampling.
This robustness results from our \modelNameCBN that modulates the pixel-wise feature during the cost regularization stage and introduces additional modulation parameters for invalid pixels (shown in Eq.~\eqref{eq:cocovono}). 
Overall, this experiment validates that our \modelNameCBN can function well under varying or non-stable LiDAR density.
}
}

\begin{figure}[t!]
    \centering
    \includegraphics[width=\linewidth]{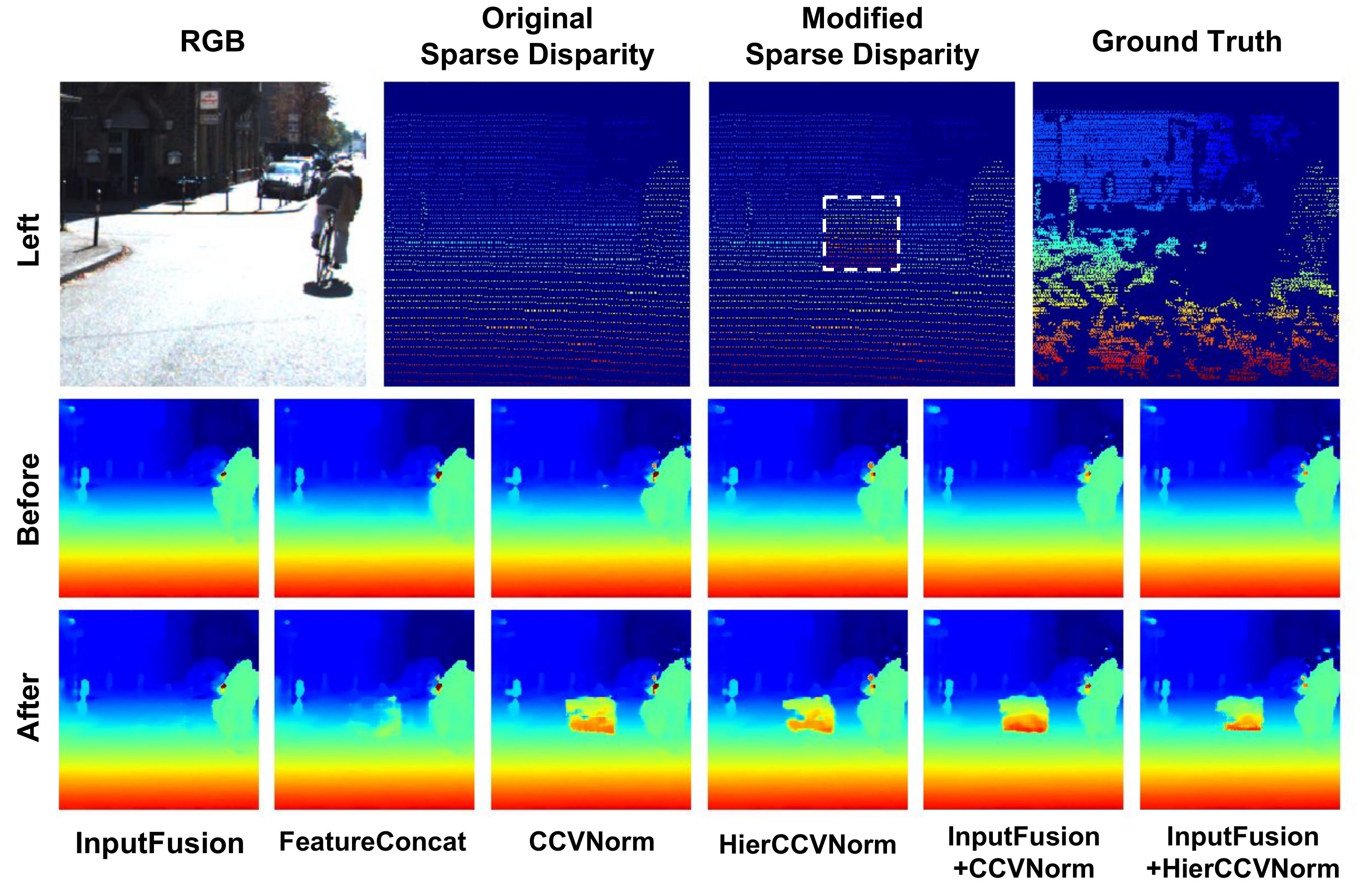}
    \caption{\small{Sensitivity to LiDAR data. We manually modify the sparse disparity input (indicated by the white dashed box in ``Modified Sparse Disparity'') and observe the effect in disparity estimates. The results show that all our variants better reflect the modification of LiDAR data during the matching process.}}
    \label{fig:vis-sensitivity}
\end{figure}

\subsection{Discussions}{

\walon{
{\flushleft {\bf Sensitivity to Sensory Inputs.}} 
In \figref{fig:vis-sensitivity}, we present an example to investigate the sensitivity of different fusion mechanisms with respect to the conditional 3D LiDAR data: we manually modify a certain portion of sparse LiDAR disparity map (indicated by the white dashed box on the third image in the top row of \figref{fig:vis-sensitivity}), and visualize the changes in stereo matching outputs produced by this modification (referring to the bottom two rows of \figref{fig:vis-sensitivity}). 
% In \figref{fig:vis-sensitivity}, we demonstrate the sensitivity and conditioning power of our \modelNameCBN and other baselines. 
% The second row is the disparity estimates given RGB and the original sparse disparity, and the third row is the prediction given RGB and the modified sparse disparity. 
% The center-located square (indicated in the white dashed box) in the modified sparse disparity map is manually multiplied by a constant factor to examine the correlation between the changes in input and output space.

Interestingly, using ``Input Fusion only'' is unaware of the modification in the LiDAR data and produces almost identical output (before v.s. after in \figref{fig:vis-sensitivity}). The reason is that fusion solely on the input level is likely to lose the LiDAR information through the procedure of network inference. For ``Feature Concat'', where the fusion is performed in the later cost regularization stage, the change starts to be visible but not significant. On the contrary, all our variants based on \modelNameCBN (or having combination with Input Fusion) successfully reflect the modification of the sparse LiDAR data onto the disparity estimation output. Hence, this verifies again our contribution in proposing proper mechanisms for incorporating sparse LiDAR information with dense stereo matching.
% In \figref{fig:vis-sensitivity}, we demonstrate the sensitivity and conditioning power of our \modelNameCBN and other baselines. 
% The second row is the disparity estimates given RGB and the original sparse disparity, and the third row is the prediction given RGB and the modified sparse disparity. 
% The center-located square (indicated in the white dashed box) in the modified sparse disparity map is manually multiplied by a constant factor to examine the correlation between the changes in input and output space. 
% Input Fusion is almost unaware of the modification in the conditioning LiDAR data since fusing in a very early stage is prone to information loss throughout network inference. 
% Feature Concat, which is operated in later cost regularization stage, still produces outputs with small changes. Meanwhile, all our variants successfully reflect the value change of the sparse LiDAR data. This suggests that our \modelNameCBN is a powerful conditioning mechanism to incorporate sparse but accurate LiDAR information with dense stereo matching.

{\flushleft {\bf Qualitative Results.}} 
\figref{fig:qualitative} provides an example to illustrate qualitative comparisons between several baselines and the variants of our proposed method. 
Our full model (i.e., \modelNameIncat + hierarchical \modelNameCBNPunc) is able to handle scenarios with complex structure by taking advantage of the complementary nature of stereo and LiDAR sensors. For instance, as indicated by the white dashed bounding box in \figref{fig:qualitative}, GC-Net fails to estimate disparity accurately on the objects containing the deformed shape (e.g., bicycles) in low illumination.
% In contrast, based on the guidance from LiDAR data which provides sparse but accurate depth sensing,
In contrast, our method is capable of capturing the details of bicycles in disparity estimation due to the help from the sparse LiDAR data.
}

%\paragraph{Sensitivity to sensory inputs.} \hubert{Can we mention about safety (more context is in comment)?}
% [Hubert] Can we mention about safety?
% Imagery input may face critical problem while dealing with extreme conditions, such as low or high lighting. In these cases, sensory inputs are more reliable and should be seriously take into account. The two critical car accident from autonomous driving are both lighting problems (one is too dark, while the other one is backlighting). 
    
\begin{figure*}[t!]
    \centering
    \includegraphics[width=\linewidth]{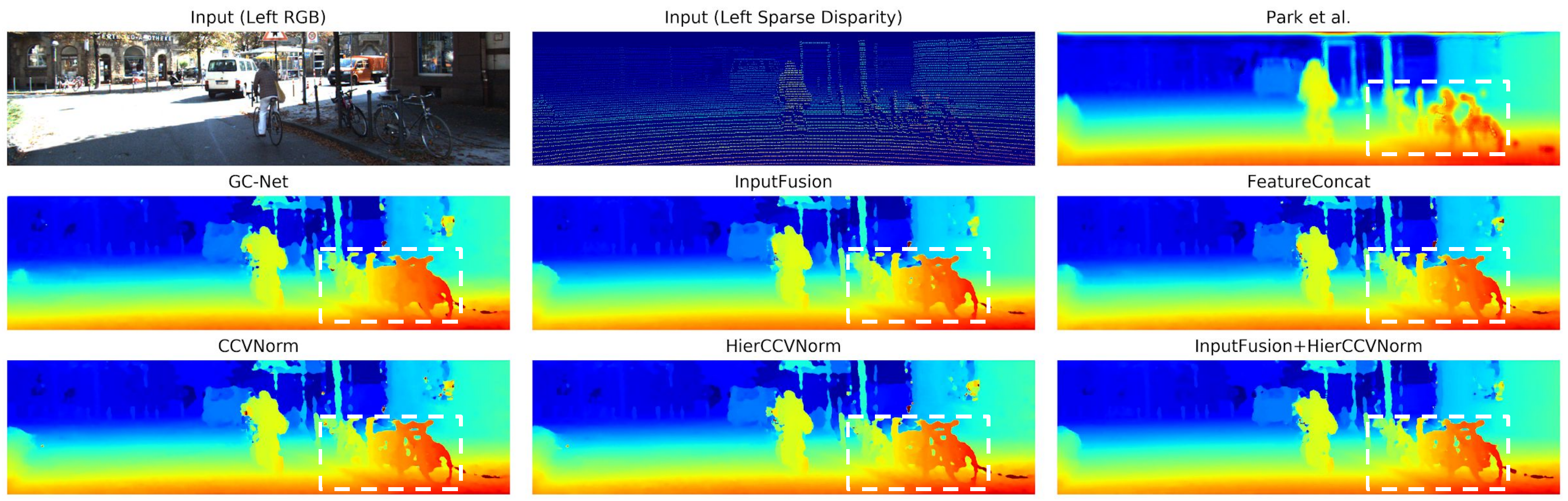}
    \caption{\small{Qualitative Results. 
    Comparing to other baselines and variants, our method captures details in complex structure area (the white dashed bounding box) by leveraging complementary characteristics of LiDAR and stereo modalities.}}
    \label{fig:qualitative}
\end{figure*}
}

\subsection{Computational Time}{
\begin{table}[!t]
\centering
\caption{\small{Computational time (unit: second). Our method only brings small overhead (0.049 seconds) compared to the baseline GC-Net.}}
\begin{tabular}{l c c c c c}
\hline
Method & SGM~\cite{stereosgm} & Prob.~\cite{LiDARstereoprobfusion} & Park.~\cite{LiDARstereoicra18} & GC-Net~\cite{stereogcnet} & Ours \\ \hline \hline
Time & $0.040$ & $0.024$ & $0.043$ & $0.962$ & $1.011$ \\
\hline
\end{tabular}
\small
\label{tab:infer-time}
\vspace{-5mm}
\end{table}

\walon{
% In \tabref{tab:infer-time}, we measure the computation time of our method and other baselines. 
We provide an analysis of computational time in \tabref{tab:infer-time}.
Except for Probabilistic Fusion~\cite{LiDARstereoprobfusion} which is tested on a Core i7 processor and an AMD Radeon R9 295x2 GPU as reported in the original paper, all the other methods run on a machine with a Core i7 processor and an NVIDIA 1080Ti GPU.
% i love dennis
In general, the models based on stereo matching networks (i.e., GC-Net and ours) take longer for computation but provide significant improvement in performance (see \tabref{tab:kitti_stereo}) in comparison with conventional algorithms.
% Compared our method with GC-Net (that our model is based on), the computational overhead introduced by our \modelNameIncat and \modelNameCBN mechanisms is marginal (0.049 second).
%
While improving the overall runtime performance via introducing more efficient stereo matching networks is out of the scope of this paper, we show that the overhead introduced by our \modelNameIncat and \modelNameCBN mechanisms upon the GC-Net method is only 0.049 seconds, validating the efficiency of our fusion scheme.
%
% suggesting that our method is a strong yet time-efficient integration of LiDAR data and stereo matching networks. 
% By the comparison between GC-Net and ours, despite the tremendous improvement in performance, stereo matching networks have its limitation in high computation time in comparison to traditional algorithms. 
% On the other hand, the computation overhead introduced by our \modelNameIncat and \modelNameCBN is trivial ($\approx 0.049$), \hubert{suggesting} our approach \hubert{is} a strong yet time-efficient integration of LiDAR data and stereo matching networks.
}

}
\section{Conclusions}

\johnson{
In this paper, built upon deep learning-based stereo matching, we present two techniques for incorporating LiDAR information with stereo matching networks: (1) \modelNameIncat that jointly reasons about geometry information extracted from LiDAR data in the matching cost computation stage and (2) \modelNameCBN that conditionally modulates cost volume feature in the cost regularization stage. Furthermore, with the hierarchical extension of \modelNameCBNPunc, the proposed method only brings marginal overhead to stereo matching networks in runtime and memory consumption. We demonstrate the efficacy of our method on both the KITTI Stereo and Depth Completion datasets. In addition, a series of ablation studies validate our method over different fusion strategies in terms of performance and robustness. We believe that the detailed analysis and discussions provided in this paper could become an important reference for future exploration on the fusion of stereo and LiDAR data.
% Finally, comprehensive visual analysis is carried out to highlight the strength of our method.
% This work opens up a new perspective to tackle stereo and LiDAR fusion with stereo matching network. Our results have shown significant improvement on performance, suggesting a potential direction for future exploration.
}

%\clearpage
\bibliographystyle{IEEEtran}
\bibliography{ref}
\normalsize

\end{document}